\documentclass[10pt,twocolumn,letterpaper]{article}

\usepackage{authblk}
\usepackage{varwidth}

\makeatletter
\renewcommand\AB@affilsepx{\hspace{1.2cm}}
\makeatother

\usepackage{iccv}

\usepackage{times}
\usepackage{amsmath}
\usepackage{amssymb}

\usepackage{multirow}

\usepackage{epsfig}
\usepackage{graphicx}
\DeclareGraphicsExtensions{.eps}
\DeclareGraphicsExtensions{.png}
\DeclareGraphicsExtensions{.jpg}

\usepackage[pagebackref=true,breaklinks=true,letterpaper=true,colorlinks,bookmarks=false]{hyperref}
\usepackage{subcaption}
\iccvfinalcopy 

\def\iccvPaperID{12} 

\ificcvfinal\fi
\begin{document}

\title{Batch-Based Activity Recognition from Egocentric Photo-Streams}

\author[1,2]{Alejandro Cartas}
\author[1,2]{Mariella Dimiccoli}
\author[1,2]{Petia Radeva}
\affil[1]{\hspace{-2.1cm}\protect\begin{varwidth}[t]{\linewidth}\protect\centering University of Barcelona\par Mathematics and Computer Science Department\par 08007 Barcelona\par Spain \authorcr {\tt\small \{alejandro.cartas, petia.ivanova\}@ub.edu} \protect\end{varwidth}}
\affil[2]{\hspace{-0.85cm}\protect\begin{varwidth}[t]{\linewidth}\protect\centering Computer Vision Center\par Universitat Aut\'onoma de Barcelona\par 08193 Cerdanyola del Vallès\par Spain \authorcr {\tt\small mariella.dimiccoli@cvc.uab.es} \protect\end{varwidth}}

\maketitle

\begin{abstract}

Activity recognition from long unstructured egocentric photo-streams has several applications in assistive technology such as health monitoring and frailty detection, just to name a few. However, one of its  main technical challenges is to deal with the low frame rate of wearable photo-cameras, which causes abrupt appearance changes between consecutive frames. In consequence, important discriminatory low-level features from motion such as optical flow cannot be estimated. In this paper, we present a batch-driven approach for training a deep learning architecture that strongly rely on Long short-term units to tackle this problem. We propose two different implementations of the same approach that process a photo-stream sequence using batches of fixed size with the goal of capturing the temporal evolution of high-level features. The main difference between these implementations is that one explicitly models consecutive batches by overlapping them. Experimental results over a public dataset acquired by three users demonstrate the validity of the proposed architectures to exploit the temporal evolution of convolutional features over time without relying on event boundaries.
\end{abstract}

\section{Introduction}
Automatic human behavior understanding has been for a long time one of the main goals of artificial intelligence practitioners \cite{pantic2006human}. Being a fundamental step towards human behavior understanding and having several application areas like healthcare, ambient intelligence, and video surveillance, activity recognition has become one of the most widely studied problems in computer vision \cite{donahue2015,ng2015,poppe2010survey,simonyan2014two,weinland2011survey}.

\begin{figure}[!t]
\newcommand\imgwidth{0.24\columnwidth}
\begin{center}
\includegraphics[width=\imgwidth]{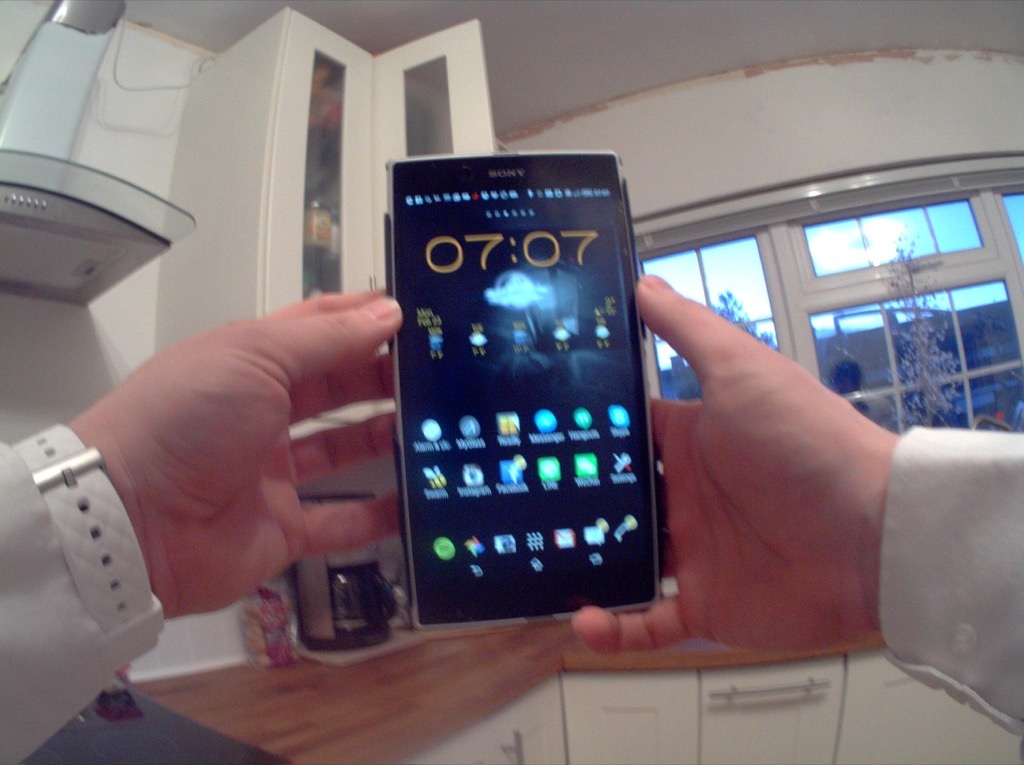}\includegraphics[width=\imgwidth]{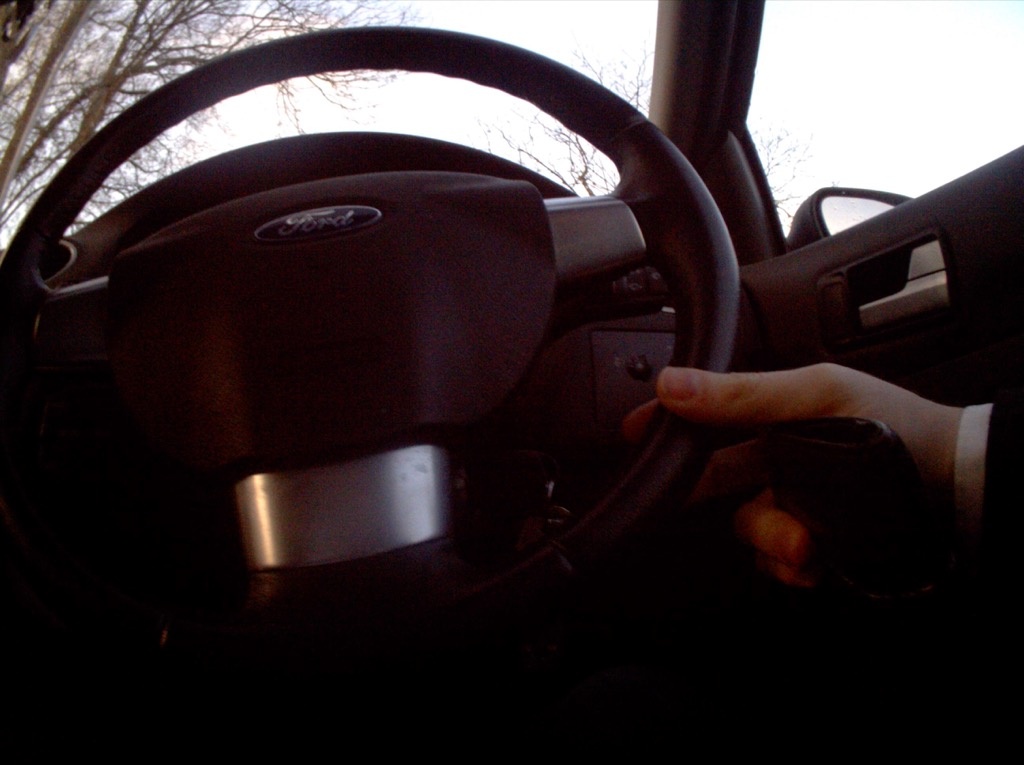}\includegraphics[width=\imgwidth]{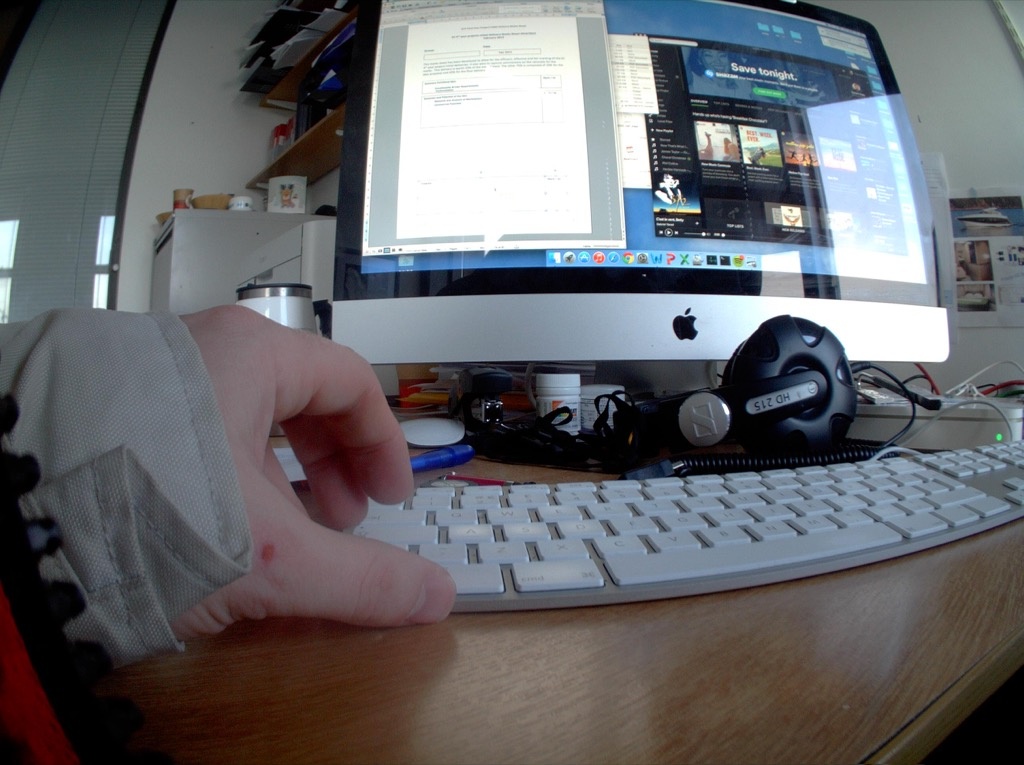}\includegraphics[width=\imgwidth]{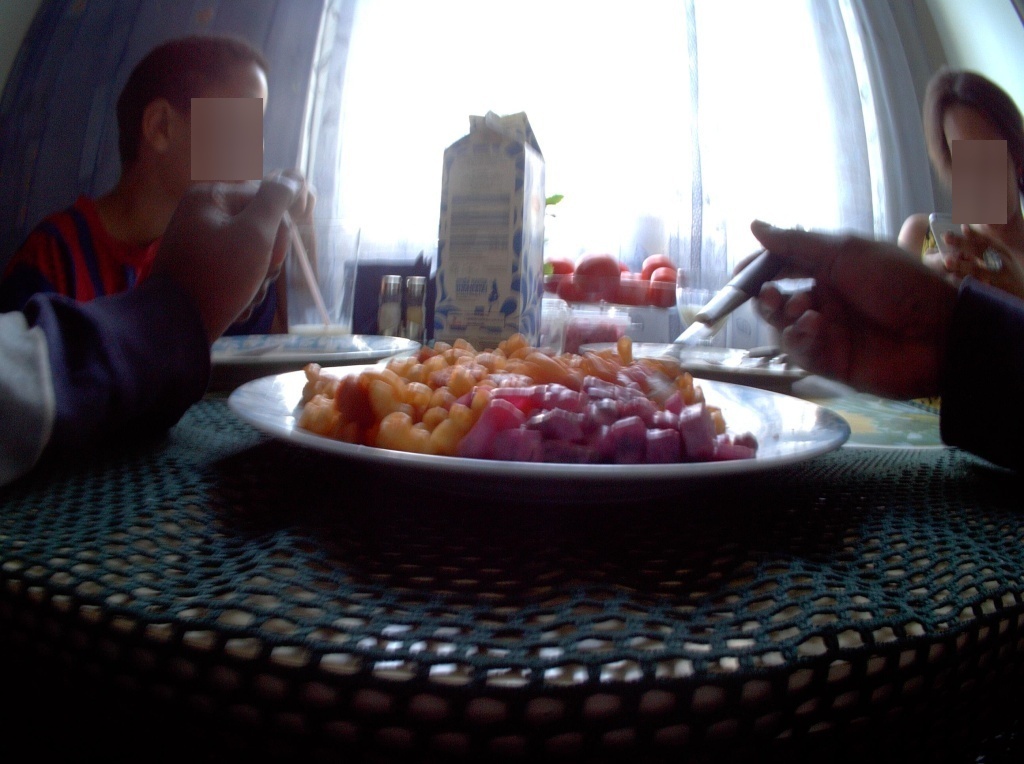}
\includegraphics[width=\imgwidth]{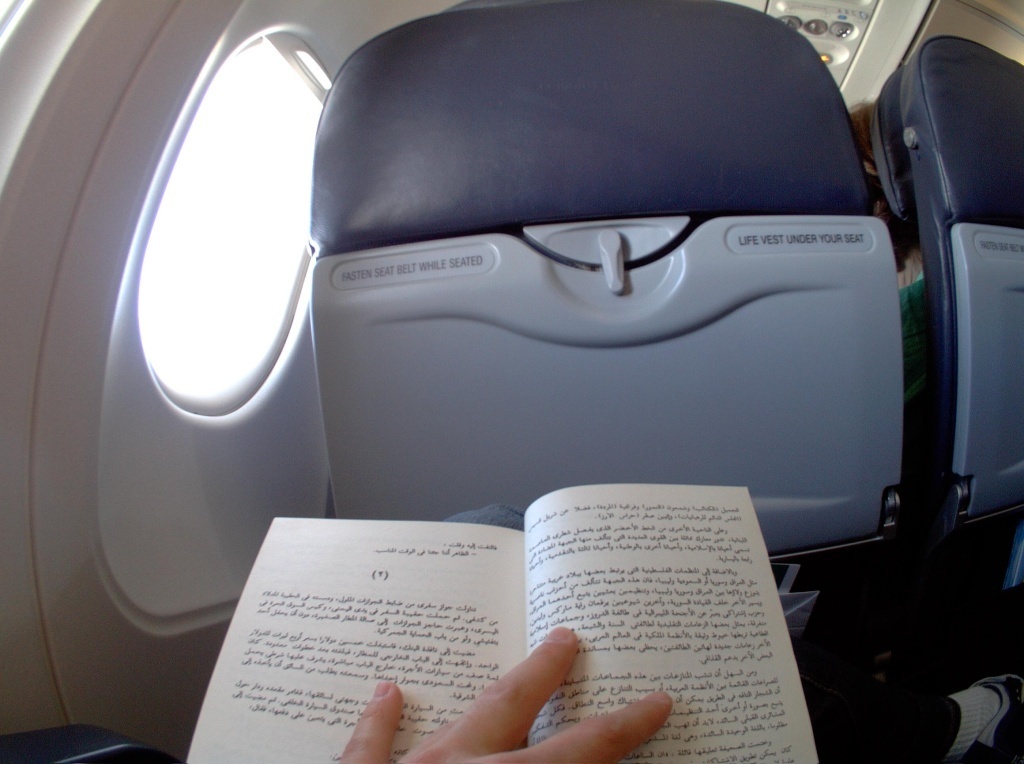}\includegraphics[width=\imgwidth]{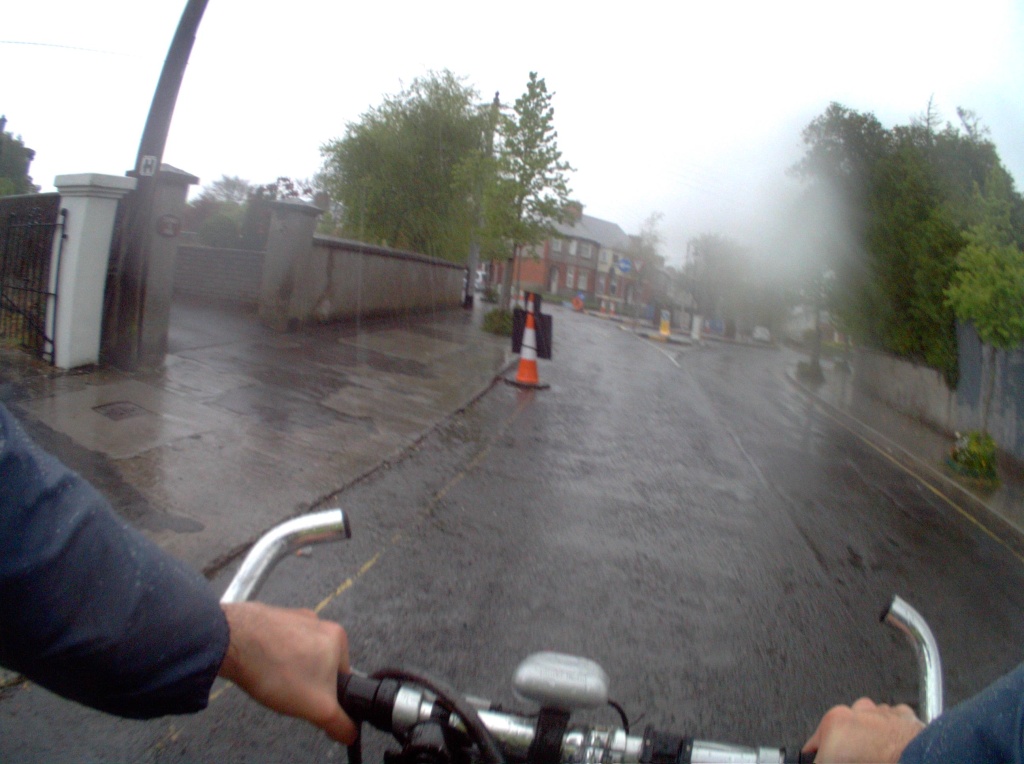}\includegraphics[width=\imgwidth]{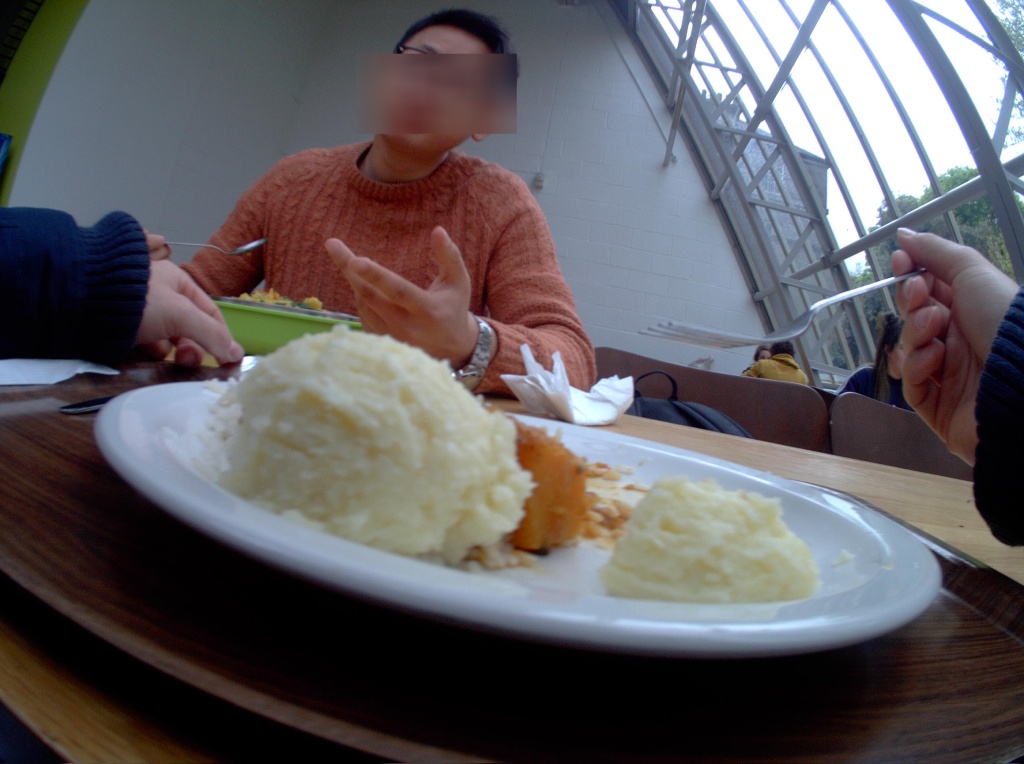}\includegraphics[width=\imgwidth]{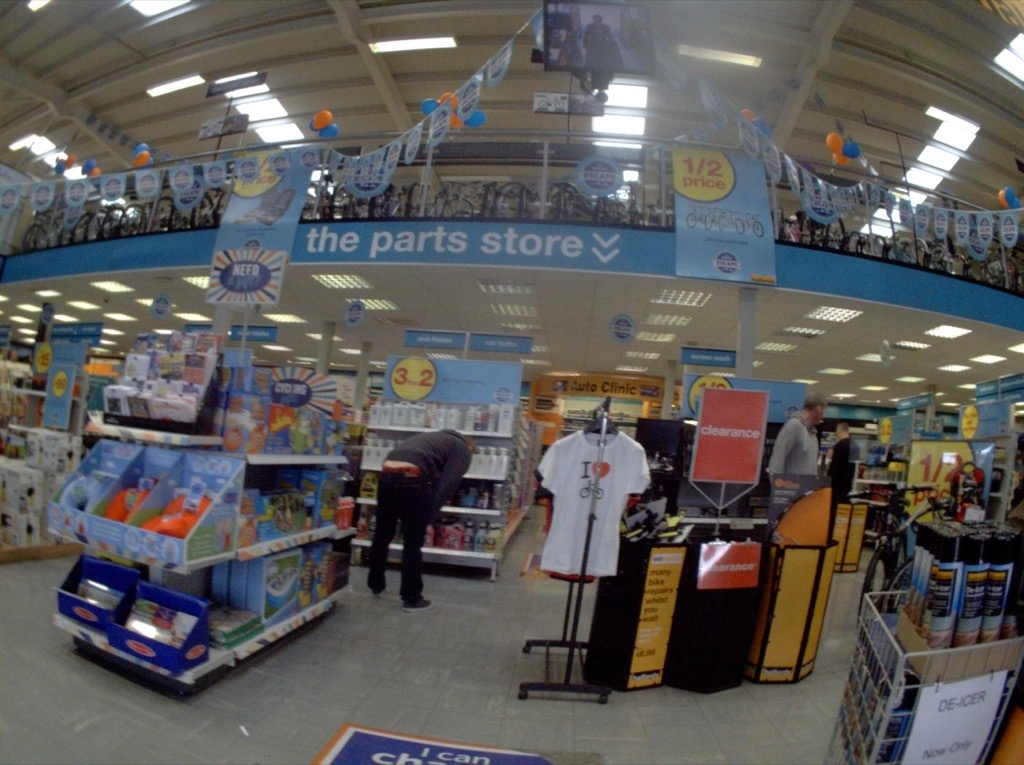}
\includegraphics[width=\imgwidth]{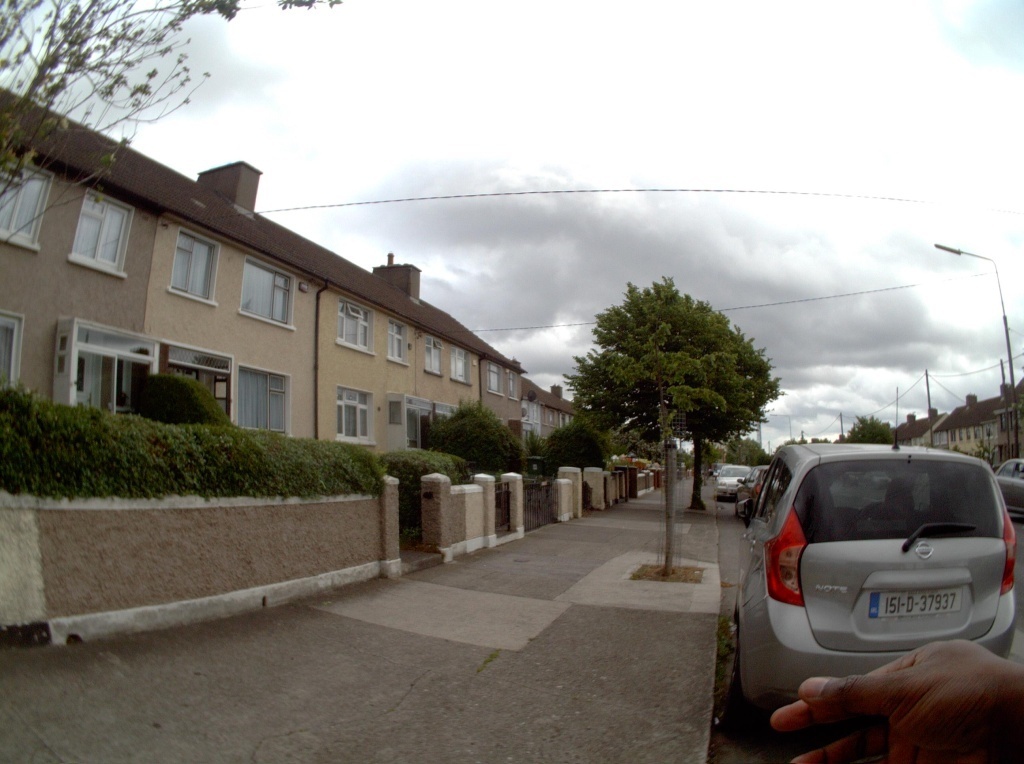}\includegraphics[width=\imgwidth]{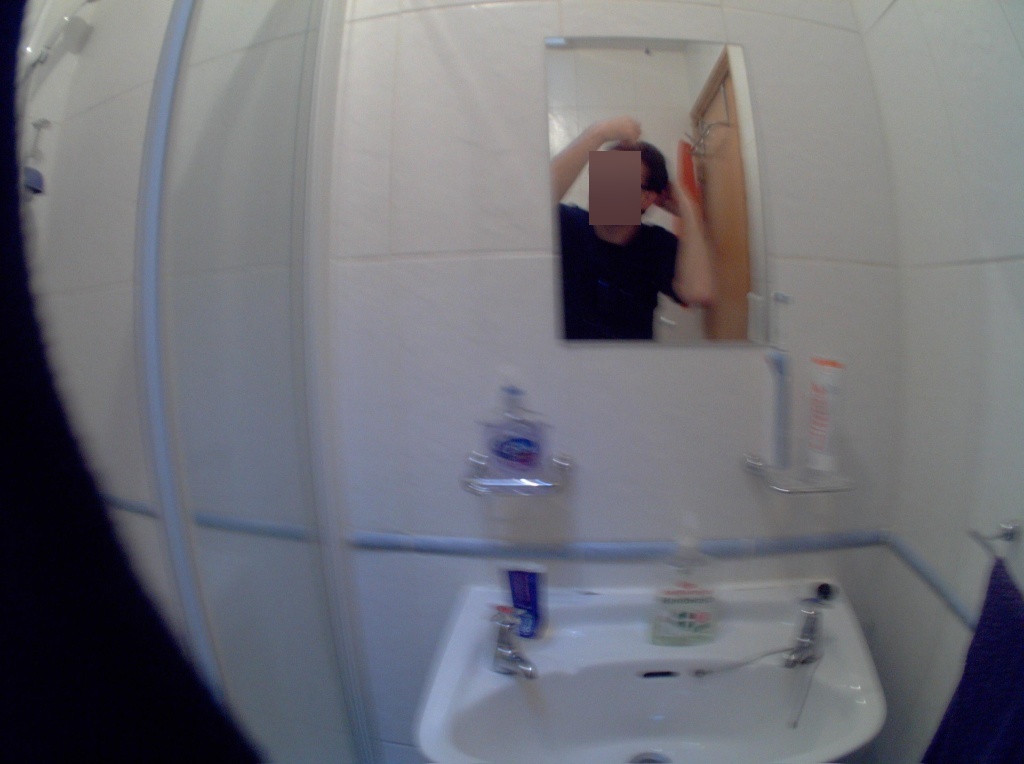}\includegraphics[width=\imgwidth]{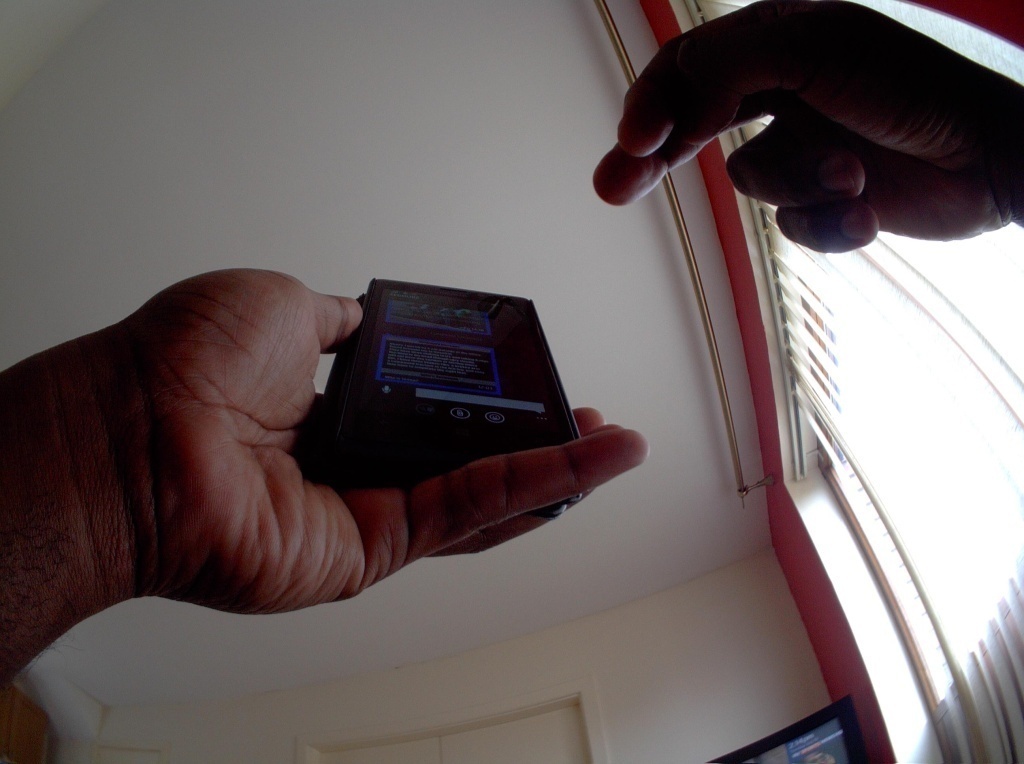}\includegraphics[width=\imgwidth]{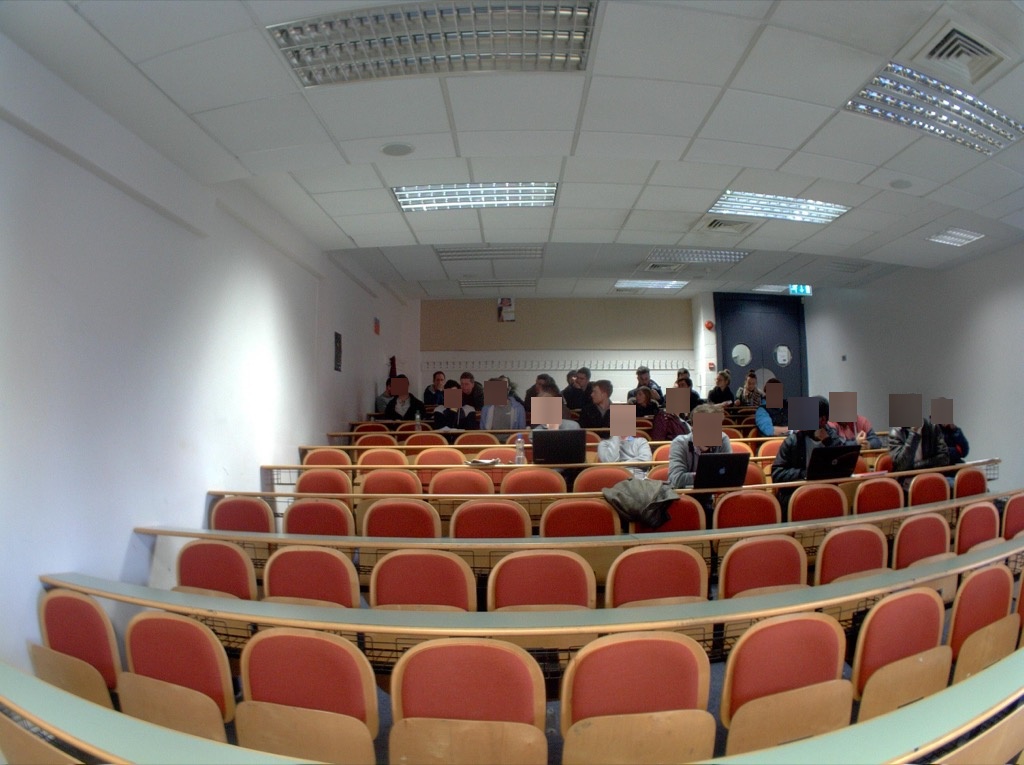}
\caption[]{Examples of egocentric images captured by a chest-mounted OMG Autographer wearable camera from the NTCIR-12 dataset ~\cite{gurrin2016NTCIR}.}
\label{fig:egocentricImage}
\end{center}
\end{figure}

With the widespread of wearable sensors in recent years \cite{guler2016brief,sazonov2014wearable}, there has been growing interest in recognizing activities from images and videos captured by a wearable camera \cite{Nguyen2016}. Since wearable cameras do not require any user intervention, they allow to capture genuine images and videos in a naturalistic setting. Additionally, the first-person point of view is specially well-suited to capture interactions with objects and people; hence tracking the activities of the wearer (see Fig. \ref{fig:egocentricImage}). However, in comparison with third-person videos, the camera free motion, the unconstrained nature of the videos, and the non-visibility of the main actor impose additional challenges to the activity recognition problem.

\begin{figure*}[!t]
\centering
\begin{minipage}{.44\textwidth}
    \centering
    \includegraphics[width=\textwidth]{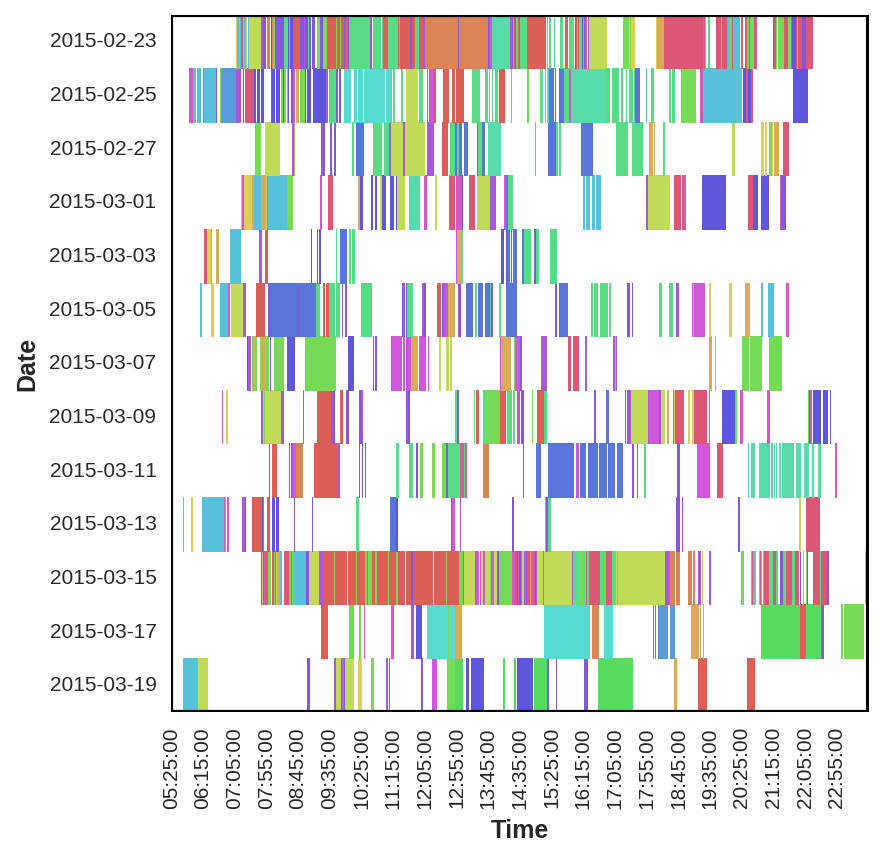}
\end{minipage}%
\begin{minipage}{.44\textwidth}
    \centering
    \includegraphics[width=\textwidth]{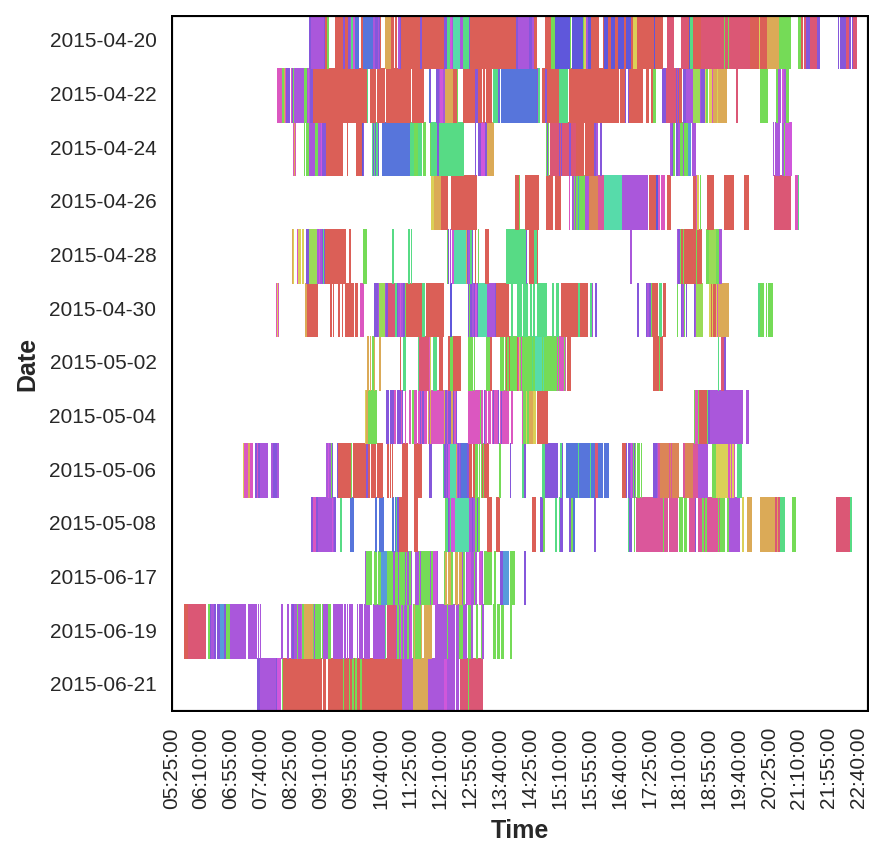}
\end{minipage}%
\begin{minipage}{.1\textwidth}
    \centering
    \includegraphics[scale=.55]{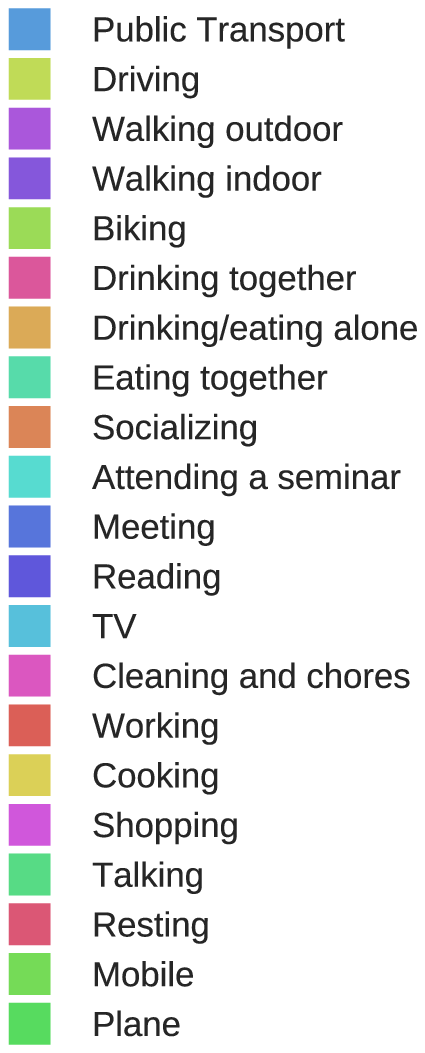}
\end{minipage}%
\caption[]{Thirteen different day sequences of annotated pictures from users \textit{u1} and \textit{u2} of the NTCIR-12 dataset.}
\label{fig:sequencesSplit}
\end{figure*}

In a lifelogging scenario, where typically the frame-rate of the camera is very low (2-3 frames per minute), the lack of temporal coherence and the abrupt changes of the field of view, further harden the activity recognition task. As stated in \cite{bolanos2017toward}, visual lifelogs offer considerable potential to infer behavior patterns through activity recognition and  enable several applications in the field of technology-driven assistive healthcare, such as preventing non-communicable diseases associated with unhealthy trends and risky profiles. 
 Despite that, to the best of our knowledge, research on activity recognition in a lifelogging scenario has received comparatively little attention in the literature \cite{cartas2017recognizing,castro2015predicting,oliveira2017leveraging} with respect to the egocentric video setting \cite{abebe2016robust,bambach2015lending,behera2012egocentric,fathi2011understanding,li2015delving,Ma_2016_CVPR,mccandless2013object,pirsiavash2012detecting,poleg2016compact,ryoo2016first,ryoo2013first}.

Activity recognition from egocentric videos is mostly focused on recognizing short-term actions such as \textit{take bread} or \textit{put ketchup}, spanning around a few hundreds of frames or more long-term activities: \textit{walking} or \textit{running} that usually last for several minutes. Not surprisingly, one of the most widely exploited features in the video setting was the ego-motion

Given the low frame-rate, activity recognition from egocentric photo-streams has focused on recognizing high-level activities that may last from a few minutes to several hours such as \textit{cooking} or \textit{working} that have proved to be well characterized by ego-motion \cite{poleg2016compact} in the video scenario. Castro et al. \cite{castro2015predicting} showed that it is possible to improve the recognition performance of a fine-tuned CNN by adding a fusion ensemble method that puts together the output of the CNN with time meta-data and contextual features through a random forest. However, the validity of the proposed approach was restricted to data belonging to a single user, since the time meta-data and the contextual features cannot generalize to multiple users. In an attempt to improve the generalization capability of this model, Cartas et al. \cite{cartas2017recognizing} proposed to use the output of a fully connected layer as additional features to be used in a fusion ensemble model. However, both  works operated at image-level even if images were manually labeled in batches. This implies that the annotators used to apply temporal information in order to label certain images, and therefore the labeling of single images could have been different without taking temporal information into account. Oliveria et al. \cite{oliveira2017leveraging} exploited  the relationship between objects and activities for activity recognition purpose. However, all the above mentioned approaches treated photo-streams as an unstructured collection of unrelated images.

\begin{figure*}[!t]
\centering
\begin{subfigure}[t]{0.45\textwidth}
	\centering
    \includegraphics[scale=0.5]{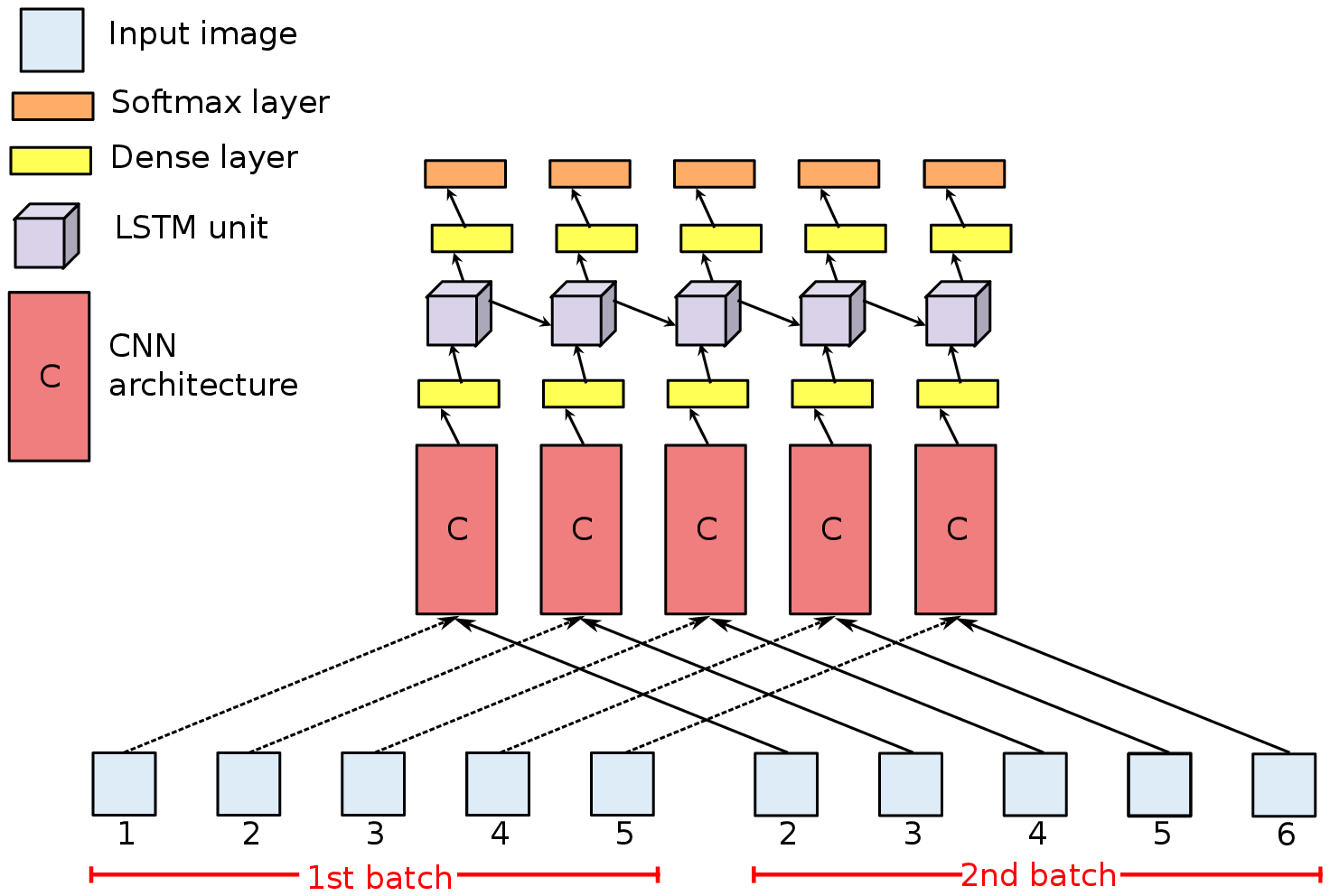}
    \caption{}
    \label{fig:architectures:1}
\end{subfigure}
\begin{subfigure}[t]{0.5\textwidth}
	\centering
    \includegraphics[scale=0.5]{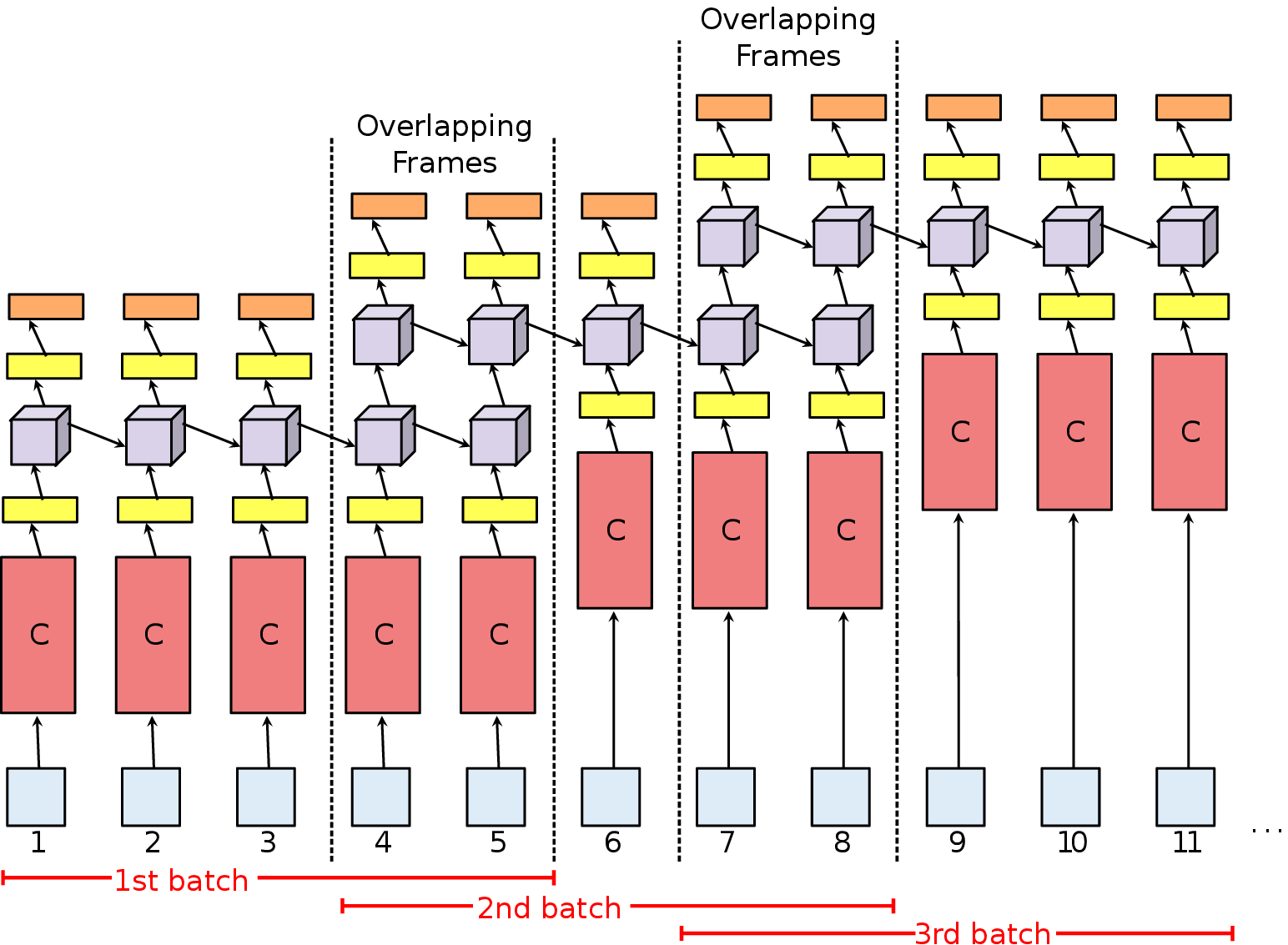}
    \caption{}
    \label{fig:architectures:2}
\end{subfigure}
\caption[]{Proposed batch-based activity recognition implementations. (a) CNN+LSTM with a timestep of 5 frames; it has 4 repeated frames between consecutive batches. (b) CNN+Piggyback LSTM with a timestep and overlap of 5 and 2 frames, respectively.}
\label{fig:architectures}
\end{figure*}

Encouraged by a previous study \cite{byrne2010everyday} showing that, besides drastic changes in appearance, the temporal coherence of concepts is preserved in egocentric photo-streams at event-level, we aim to investigate how to take advantage of temporal information to improve the recognition performance. 
Our proposed approach is similar to \cite{yue2015beyond}, where the activity recognition problem from third-person cameras is  cast as a video classification problem, whose goal is to learn a global description of the video while maintaining a low computational cost. In \cite{yue2015beyond}, the video is down-sampled to a frame-rate of 1 fpm, and explicit motion information in the form of optical flow images computed over adjacent frames is added to compensate the lost of implicit motion information. A Long Short Term Memory (LSTM) recurrent neural network operating on frame-level CNN activations is used to discover long-range temporal relationships and to learn how to integrate information over time. Contrary to \cite{yue2015beyond}, in our lifelogging scenario, image sequences have been originally recorded with a low frame-rate, so that motion information is not available. Furthermore, since a single video for us corresponds to the set of all images of the day, the number of labels may be arbitrarily large and activity boundaries are unknown. Therefore, we propose a batch-based deep learning training approach aiming to cope with both the lack of knowledge about event boundaries and the not negligible length of photo-streams (up to 2,000 images).

Our contributions can be summarized as follows:
\begin{itemize}
\item We propose two end-to-end batch-based implementations that exploit the temporal coherence of concepts in photo-streams and outperform the state of the art end-to-end architectures.
\item We demonstrate that it is possible to capture the temporal evolution of features over time in photo-streams even without knowing event boundaries.
\item We show that both implementations improve the classification performance, but the first one is slightly better on day sequences that does present clear temporal patterns.
\end{itemize}

The rest of the paper is organized as follows: section \ref{sec:approach} details the proposed approach, whereas the experimental setting, validations and state-of-the-art comparisons are described and discussed in section \ref{sec:experiments}. Concluding remarks are reported in section \ref{sec:conclusions}.

\begin{figure*}[!t]
\begin{center}
\includegraphics[width=\textwidth]{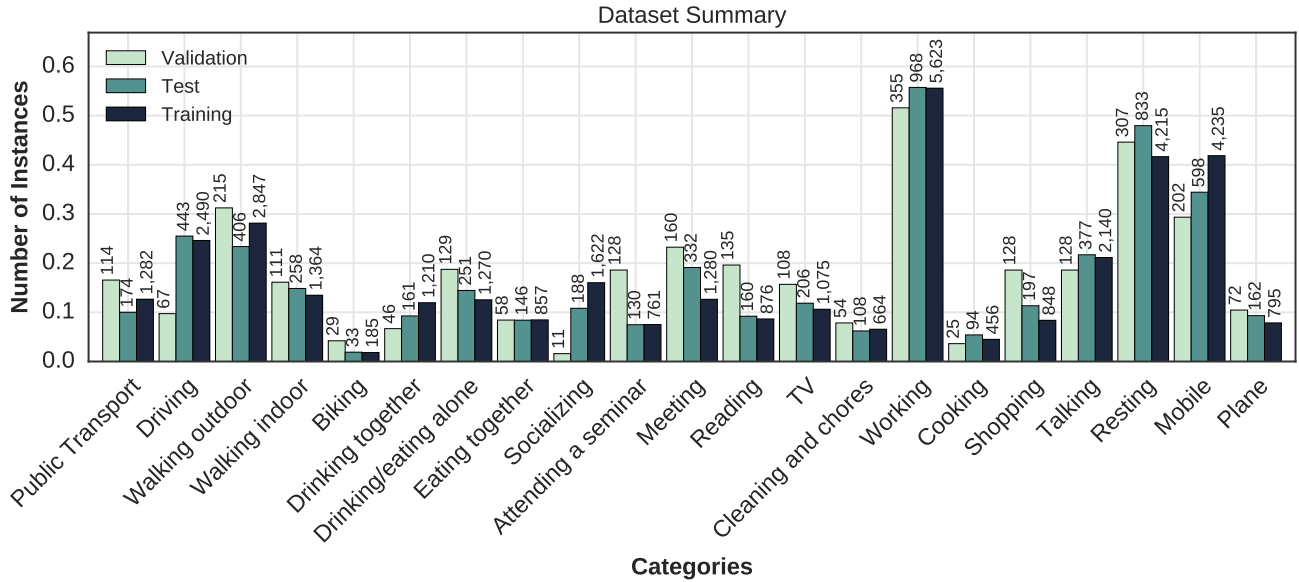}
\caption[]{Dataset split summary. All split distributions are normalized, but the corresponding number of instances for each category is shown on top of each split.}
\label{fig:datasetSummary}
\end{center}
\end{figure*}

\section{Proposed approach}
\label{sec:approach}

Our hypothesis is that the variations between successive images in a photo-stream encode additional information which could be useful in making more accurate activity predictions. While this hypothesis has been validated in \cite{yue2015beyond} for conventional videos whose frames share the same label, there is no evidence that it is applicable to the photo-stream scenario. Typically, a photo-stream  has several labels corresponding to different activities performed by the user during a day. Due to the sparseness of the observations, adjacent frames may have distinct labels even if in general, when the user is performing a long term activity, several consecutive frames share the same label (see Figure \ref{fig:sequencesSplit}). One possible approach would be to first split the photo-stream or video into events manually or by using a state of the art approach as \cite{talavera2015r,dimiccoli2016sr}, and then to classify each event separately. However,  we propose two different CNN-LSTM implementations,based on the idea of batch-based training,  able to cope with the whole photo-stream set analysis without the necessity for event segmentation.

In the first implementation (see Fig. \ref{fig:architectures:1}), the output from the CNN layer is given as input to a single LSTM layer. While the architecture itself is not new, the way it is trained and what it is supposed to learn differs from \cite{yue2015beyond}. During training, we split the photo-stream into overlapping segments of fixed length and we feed them, together with the corresponding activity label at frame level, to the network. Since each segment has a fixed length, its images may not share the same label. Therefore, we expect the network to learn not only the temporal evolution of features over time within a same event, but also to learn to predict event changes, thus leading to more accurate predictions when event boundaries are unknown.

In the second implementation  (see Fig. \ref{fig:architectures:2}), we explicitly model the temporal relation between two adjacent overlapping batches.  After a CNN architecture, we added a \textit{fully-connected} layer, a LSTM unit, and one last \textit{fully-connected} layer. For frames belonging to the first batch of a sequence (frames 1-5 in Fig. \ref{fig:architectures:2}(b)),  the input of the LSTM layer is the output of the \textit{fully-connected} layers. For frames belonging to two temporally overlapping batches (frames 4-5 in Fig. \ref{fig:architectures:2}(b)), the input of the LSTM layer is the output of the  LSTM layer from the previous batch. The outputs of the first \textit{fully-connected} layer and the LSTM unit have the same size. In this way, the input of the LSTM layer has the same size of its output, that in the latter is used in subsequent passes. In other words, after the initial feedfoward pass of the first sequence batch, the LSTMs output of the last $n$ frames are stored. In subsequent passes, they are used as the input for the first $n$ LSTMs. For example, Fig. \ref{fig:architectures:2} shows this configuration composed of 5 timesteps with an overlapping of 2 output/input. This architecture is supposed to learn more complex long-range temporal dependencies without the need of considering each one-day photo-stream as a single sequence. Indeed, since a photo-stream can be made of up to two thousands frames, even for an LSTM it would be unfeasible to learn such long range dependencies\cite{gers2000learning}.
Using both implementations, we finally obtain the frame-level predictions.

\begin{figure*}[!t]
\centering
\begin{subfigure}[t]{0.33\textwidth}
    \hspace{-0.9cm}\includegraphics[scale=0.6]{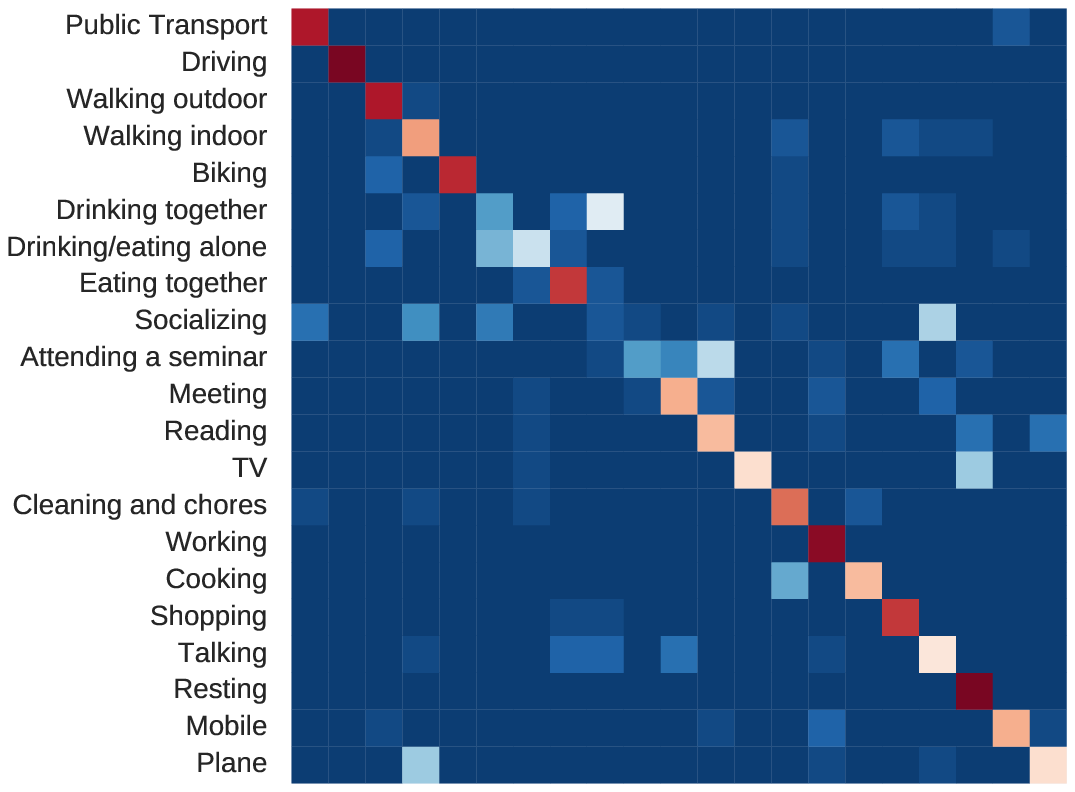}
    \caption{VGG-16}
    \label{fig:confusionMatrices:1}
\end{subfigure}
\begin{subfigure}[t]{0.27\textwidth}
    \centering
    \includegraphics[scale=0.615]{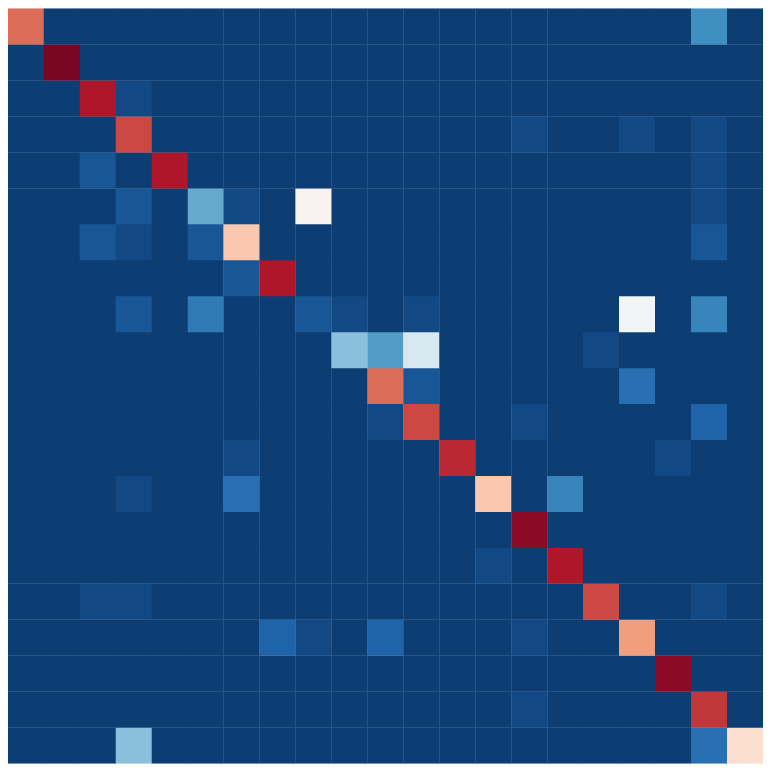}
    \caption{VGG-16+LSTM\\
(timestep of 15)}
    \label{fig:confusionMatrices:2}
\end{subfigure}
\begin{subfigure}[t]{0.3\textwidth}
    \centering
    \includegraphics[scale=0.6]{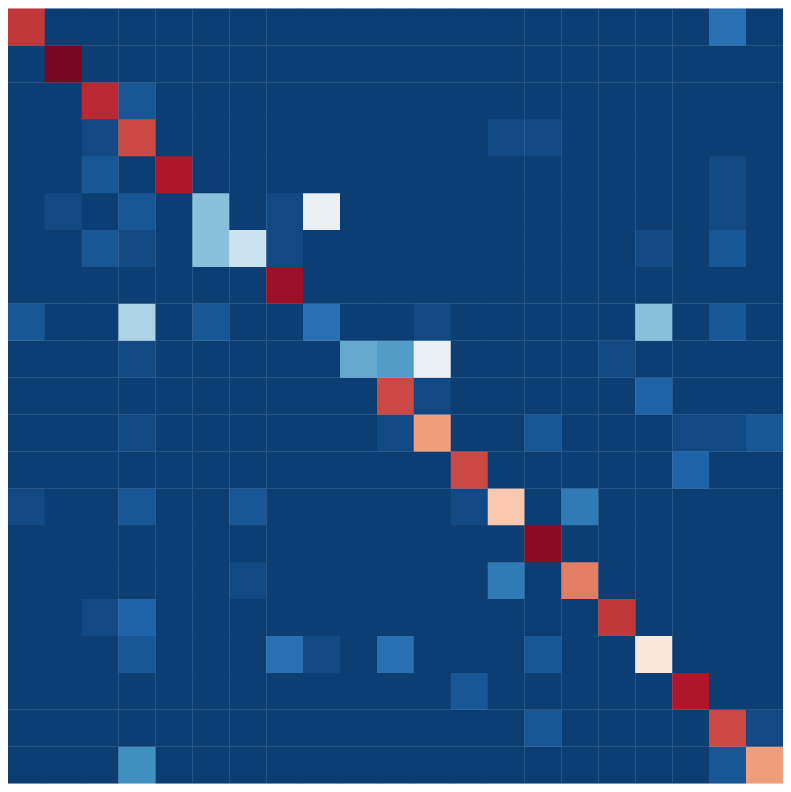}
    \caption{VGG-16+Piggyback LSTM\\
(timestep of 10 and overlap of 3)}
    \label{fig:confusionMatrices:3}
\end{subfigure}
\begin{minipage}[b]{.05\textwidth}
\includegraphics[scale=0.6]{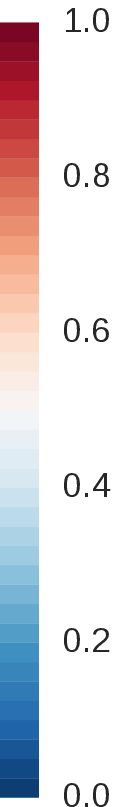}
\end{minipage}
\caption[]{Normalized confusion matrices of all models for the best trained configuration.}
\label{fig:confusionMatrices}
\end{figure*}

\section{Experiments}
\label{sec:experiments}

We describe the dataset in section \ref{sec:datasets} and detail the networks training in section \ref{sec:training}. We then present the experimental results on activity recognition on section \ref{sec:results}.

\subsection{Dataset}
\label{sec:datasets}

We employed the NTCIR-12 dataset~\cite{gurrin2016NTCIR} on our experiments. This dataset contains 89,593 egocentric images collected in 79 days by three different persons. The data collection was done in a period of almost a month per person. During this time, each user worn a chest-mounted camera that took two pictures per minute. Continuing previous work~\cite{cartas2017recognizing}, we used an extended subset of 44,902 images from the NTCIR-12 dataset, around 15,000 images per person. These images were annotated using 21 activity categories and correspond to all three users and 78 days at different times. The annotation process was done in batches of consecutive frames, meaning that the context of a continuous activity across frames was implicitly taken into account by the annotators.

We split the annotated images subset in training, validation, and test sets. These splits contain full day sequences and maintain the inherit class imbalance, as illustrated on Fig. \ref{fig:datasetSummary}. We accomplished this by doing the following. First, the day sequences were grouped in bins of similar number of images by using the first-fit decreasing algorithm. Second, all possible combinations of test splits from the bins were calculated by using the Twiddle algorithm~\cite{Chase1970}. Then, two category distributions were computed for each test split combination and its remaining bins. Then, for each pair of distributions the sum of the Bhattacharya distances between them and the whole dataset category distribution was obtained. Finally, the best test split is the one with the shortest distance. The validation and training split was calculated on the remaining bins by doing the same steps.

For the recurrent proposed models, the annotated pictures of a day were considered as one sequence. In total, the training and validation set consisted of 59 and 7 sequences, respectively. For example, thirteen day sequences from two users are shown in Fig. \ref{fig:sequencesSplit}.

\subsection{Training}
\label{sec:training}

All the models were implemented using the Keras framework~\cite{chollet2015keras}. In addition, all the models have the same data augmentation process at the frame level. Namely, we randomly applied horizontal flips, translation and rotation shifts, and zoom operations. To avoid overfitting, we added dropout layers~\cite{srivastava14a} and weight normalization to all models. The VGG CNN architecture is used to process individual images. To aggregate photostream-level information, we leverage an LSTM to consider sequences of CNN activations. The training and configuration details for all models are described as follows.

\textbf{VGG-16 CNN}. We fine-tuned a VGG-16 network~\cite{Simonyan14c} as our base model. Only the last fully-connected layer was changed to have a 21 category output. The fine-tuning was done in two phases using the splits described in the previous section.The goal of the first phase was to initialize the weights at the top layers, since only the bottom layers of the CNN were initialized using the weights of the ImageNet classification task. Therefore, during the first phase, only the fully-convolutional layers were backpropagated. The optimization method used was the Stochastic Gradient Descent (SGD) for 10 epochs, a learning rate $\alpha=1\times10^{-5}$, a batch size of 1, a momentum $\mu=0.9$, and a weight decay equal to $5\times10{-6}$. In the second phase, the last two convolutional layers were also fine-tuned and the initial weights were obtained from the best epoch of the first phase. Moreover, the SGD ran for another 10 epochs and set with the same parameters except the learning rate $\alpha=4\times10^{-5}$. The best validation results were obtained at the eight epoch.

\begin{table}[!t]
\centering
\resizebox{1.0\columnwidth}{!}{%
\begin{tabular}{ | c | c |c |c |c |}
\hline
\multirow{2}{*}{Method} & \multirow{2}{*}{Accuracy} & Macro & Macro & Macro \\
& & Precision & Recall & F1-score \\ \hline
VGG-16 & 75.97 & 68.50 & 67.49 & 66.80 \\ \hline
VGG-16+LSTM  & \multirow{2}{*}{79.68} & \multirow{2}{*}{72.96} & \multirow{2}{*}{71.36} & \multirow{2}{*}{70.87} \\
timestep 5 & & & & \\ \hline
VGG-16+LSTM  & \multirow{2}{*}{80.39} & \multirow{2}{*}{75.25} & \multirow{2}{*}{71.86} & \multirow{2}{*}{71.97} \\
timestep 10 & & & & \\ \hline
VGG-16+LSTM  & \multirow{2}{*}{\textbf{81.73}} & \multirow{2}{*}{\textbf{76.68}} & \multirow{2}{*}{\textbf{74.04}} & \multirow{2}{*}{\textbf{74.16}} \\
timestep 15 & & & & \\ \hline
VGG-16+Piggyback LSTM  & \multirow{2}{*}{75.97} & \multirow{2}{*}{69.74} & \multirow{2}{*}{62.98} & \multirow{2}{*}{63.24} \\
timestep 5 overlap 2& & & & \\ \hline
VGG-16+Piggyback LSTM  & \multirow{2}{*}{79.04} & \multirow{2}{*}{72.98} & \multirow{2}{*}{71.88} & \multirow{2}{*}{71.06} \\
timestep 10 overlap 3& & & & \\ \hline
VGG-16+Piggyback LSTM  & \multirow{2}{*}{78.51} & \multirow{2}{*}{73.00} & \multirow{2}{*}{69.52} & \multirow{2}{*}{69.88} \\
timestep 15 overlap 4& & & & \\ \hline
\end{tabular}
}
\caption{Performance summary of all methods with different number of timesteps and overlapping frames.}
\label{tab:performanceSummary}
\end{table}

\begin{figure*}[!t]
\begin{center}
\begin{minipage}{0.84\textwidth}
\centering
\includegraphics[width=\textwidth]{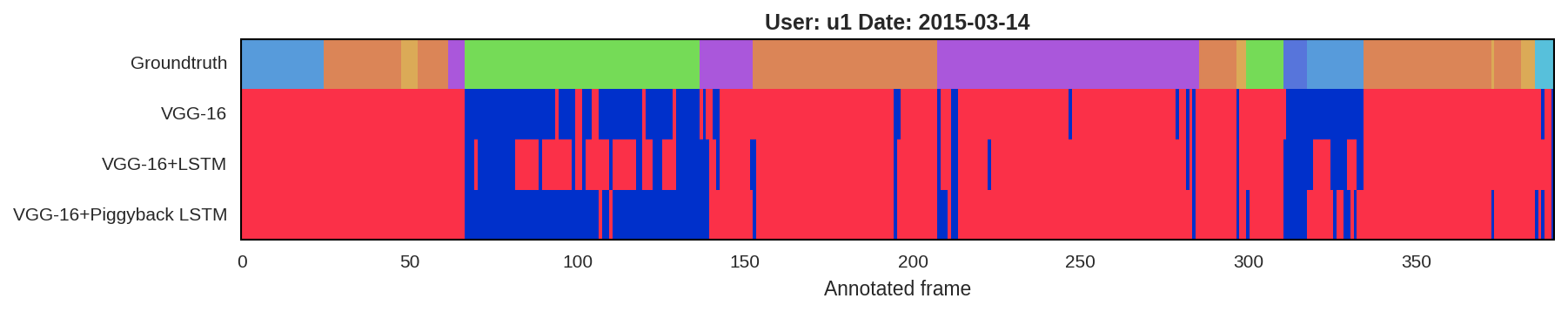}
\includegraphics[width=\textwidth]{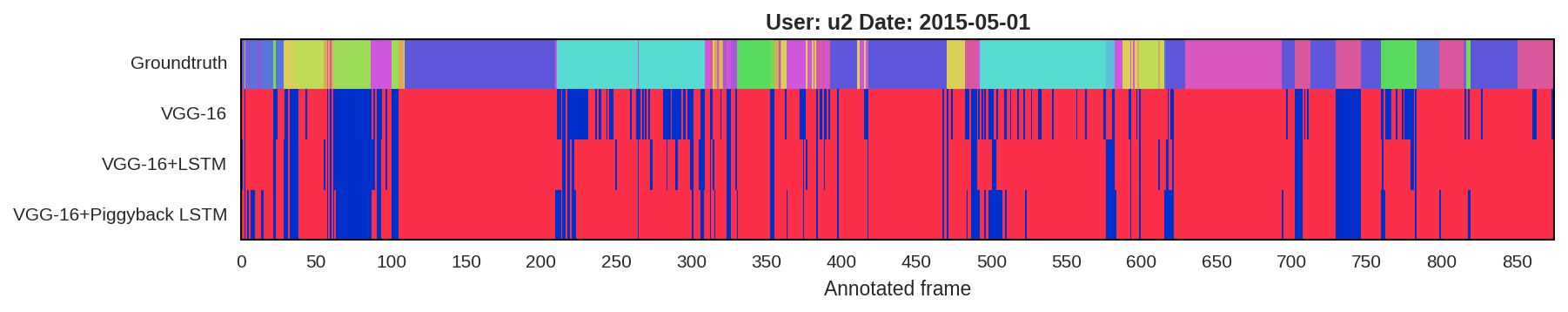}
\includegraphics[width=\textwidth]{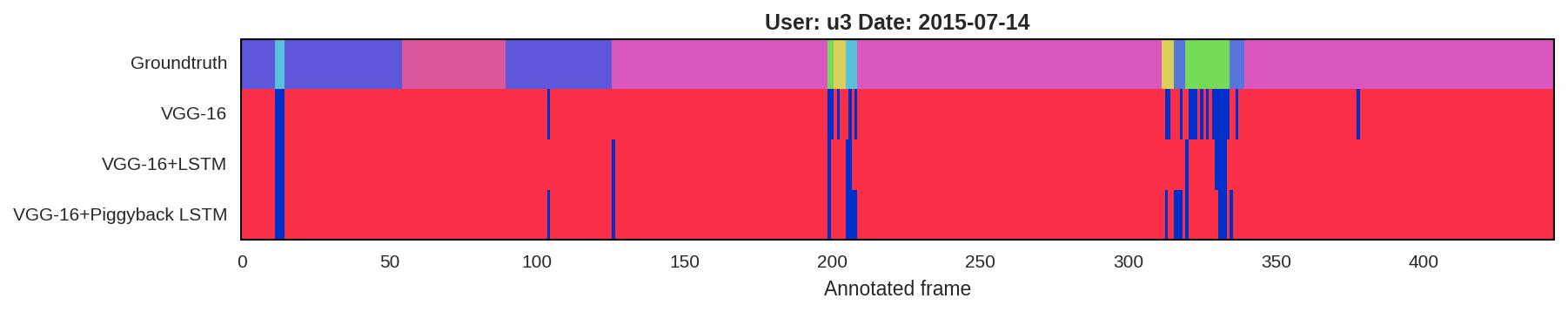}
\end{minipage}
\begin{minipage}{0.15\textwidth}
\includegraphics[scale=0.5]{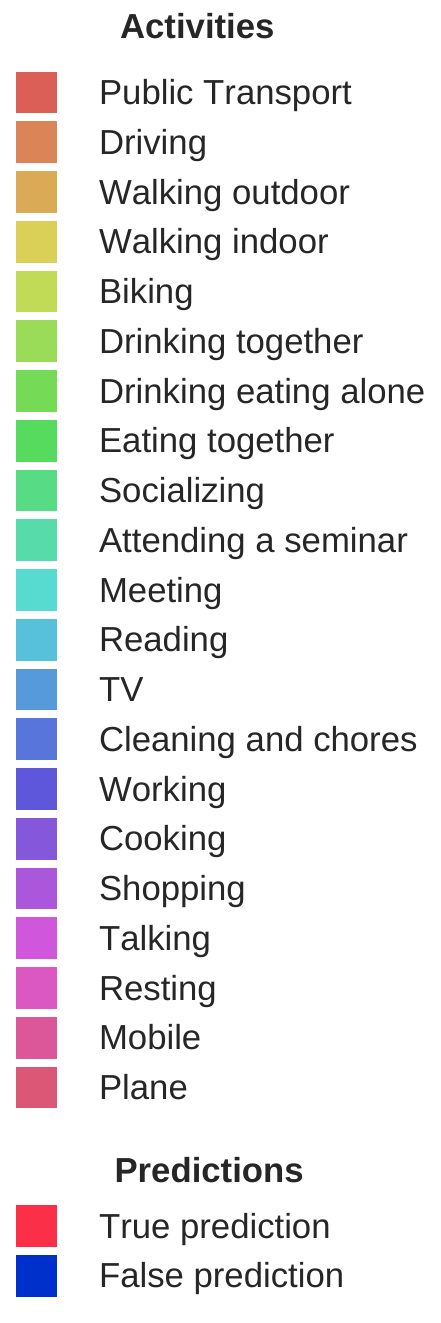}
\end{minipage}
\caption[]{Prediction comparison of all models for three test day sequences from all users in the dataset.}
\label{fig:sequencePredictions}
\end{center}
\end{figure*}

\textbf{VGG-16 CNN + LSTM}. For this architecture, we added a LSTM layer of 256 units followed by a fully-connected layer after the first fully-connected layer of the VGG-16 network. Furthermore, in our experiments, we used three LSTM configurations with a time step of five, ten, and fifteen frames. In order to train all the configurations, we froze the weights of the first four blocks of the convolutional layers. During training, we used SGD as optimization algorithm. For the timestep 5 configuration, we trained it for 5 epochs with a learning rate $\alpha=2.5\times10^{-5}$, a momentum $\mu=0.9$, and a weight decay equal to $5\times10{-6}$. For the timestep 10 configuration, we trained it for 4 epochs with a learning rate $\alpha=1\times10^{-4}$, a momentum $\mu=0.9$, and a weight decay equal to $5\times10{-6}$. The timestep 15 configuration was trained for 2 epochs with a learning rate $\alpha=1\times10^{-4}$, a momentum $\mu=0.9$, and a weight decay equal to $5\times10{-6}$. The initial weights of the optimization process were the ones obtained for the base model. The training was performed in batches of 5, 10, and 15 frames, respectively. These batches were sampled using a sliding window of frames from each sequence. For instance, Fig. \ref{fig:architectures:1} shows an unrolled version of this model and two consecutive batches of 5 frames. Besides being a crucial characteristic of the proposed approach, this form of training can be understood as a kind of data augmentation over the sequence.

\textbf{VGG-16 CNN + Piggyback LSTM}. For this architecture, we added a fully-connected and LSTM layers after the convolutional architecture of the VGG-16 network. Both layers have an output vector length of 256 and are followed by a dense layer with a softmax activation. The feedback from overlapping frames between batches was implemented using a filter layer. Therefore, the network had two additional inputs: one input for the previous batch and another one used as a mask. We used three different configurations in the reported experiments. The first configuration had a batch size of 5 frames and 2 overlapping frames, the second one had a batch size of 10 frames and 3 overlapping frames, and the last one had a batch size of 15 frames and 4 overlapping frames. Our results were achieved by dividing the training into two phases. The purpose of the first training phase was to learn the high-level features from adjacent frames, while the purpose of the second one was to learn temporal patterns from them throughout the sequence. During the first phase, the day sequences were considered as consecutive batches without overlapping. Accordingly, this phase followed the same training procedure as the previous described architecture. In the second phase, we froze all the convolutional layers and the first fully-connected layer. The data augmentation process for the day sequences consisted in the following. Given a batch size $n$ and a overlapping number of frames $m$, the sequential training batches with $m$ overlapping frames were created from the day sequences starting at the first $n$ frames. This created more training examples than the previous architecture and considered all the frames of a sequence. The first training step was to randomly shuffle the day sequences at each epoch. Then, all the day sequences were processed one by one. The ordered training batches from a day sequence were consecutively feedforwarded to the network. The SGD algorithm was also used for this second phase. The learning rates for the configurations were $\alpha=2.5\times10^{-5}$ , $\alpha=1\times10^{-4}$, and $\alpha=1\times10^{-4}$, correspondingly. Moreover, all configurations shared the same momentum $\mu=0.9$ and weight decay equal to $5\times10{-6}$. Early stopping criteria was used for both training phases.

\subsection{Results and Discussion}
\label{sec:results}

The evaluation of the models and their configurations was done over 12 day sequences from all users, i.e. 6,225 images. In contrast with previous works ~\cite{donahue2015,yue2015beyond}, we are interested in a many-to-many sequence classification and we did not apply any kind of average over a processed batch. Since the dataset is unbalanced, we used other metrics besides accuracy to measure the classification performance. Table \ref{tab:performanceSummary} shows the obtained results for the activity recognition task. These results demonstrated that processing batches of sequential frames improves accuracy performances with respect to the pure CNN baseline for most of the cases. A larger time-step is generally preferred since it allows to better capture the temporal evolution of features over time. By using the first proposed architecture, we achieve an improvement of more than $4\%$ with respect to the VGG-16 baseline even considering a very small batch size. The results of the second proposed architecture also improved all the performance metrics with respect to the pure CNN baseline, but only for the largest timestep configuration. In comparison with the former architecture, the overall performance improvement was less. This might be explained as a consequence of not having clear temporal activity patterns throughout the whole day sequences, as shown in Fig. \ref{fig:sequencesSplit}. Moreover, a comparison of the models over three day sequences is illustrated in Fig. \ref{fig:sequencePredictions}. Although, the proposed architectures improved the overall accuracy, they still failed at classifying categories highly correlated like \textit{Eating together} and \textit{Driking together} as seen on the confusion matrices on Fig. \ref{fig:confusionMatrices}.

\section{Conclusions}
\label{sec:conclusions}

We presented a batch-based learning approach for activity recognition from egocentric photo-stream sequences. In order to learn temporal activity patterns between frames, both proposed implementations of this approach uses a LSTM unit on top of a convolutional neural network to process a day sequence of frames using windows of fixed size. Specifically, our first implementation uses a sliding window of consecutive frames to generate training batches. Moreover, our second implementation is able to handle information of previous batches from a sequence by reprocessing a fixed number of overlapping frames.

Although this paper has demonstrated that it is possible to exploit temporal coherence of concepts without knowing event boundaries, we consider that clustering a day sequence into different scene subsequences could further improve the activity recognition. Additionally, we think that the second proposed implementation could improve the activity recognition performance on video data. Both ideas will be addressed in future work.

\section*{Acknowledgments}

A.C. was supported by a doctoral fellowship from the Mexican Council of Science and Technology (CONACYT) (grant-no. 366596). This work was partially founded by TIN2015-66951-C2, SGR 1219, CERCA, \textit{ICREA Academia'2014} and 20141510 (Marat\'{o} TV3). The funders had no role in the study design, data collection, analysis, and preparation of the manuscript. M.D. is grateful to the NVIDIA donation program for its support with GPU card.

{\small
\bibliographystyle{ieee}

}
\end{document}